\documentclass[letterpaper, 10 pt, conference]{ieeeconf}  

\IEEEoverridecommandlockouts                              

\usepackage{xcolor}
\usepackage{algorithm}
\usepackage{amssymb}
\usepackage{algpseudocode}
\usepackage{graphicx}
\usepackage{soul}

\usepackage{amsmath} 

\title{\LARGE \bf
Realtime Robust Shape Estimation of Deformable Linear Object
}

\author{Jiaming Zhang$^{1,2}$, Zhaomeng Zhang$^{2}$, Yihao Liu$^{1,2,\dagger}$, Yaqian Chen$^{2}$, \\Amir Kheradmand$^{3,4,\ddagger}$, Mehran Armand$^{1,2,5,\ddagger}$
\thanks{This work was partially supported by a grant from the National Institute of Deafness and Other Communication Disorders (R01DC018815).}
\thanks{$^{1}$ Department of Computer Science, The Johns Hopkins University, Baltimore, MD, USA
        {\tt\small (jzhan282@jhu.edu, yliu333@jhu.edu, mehran.armand@jhuapl.edu)}}%
\thanks{$^{2}$ Biomechanical- and Image-Guided Surgical Systems (BIGSS) laboratory within LCSR, The Johns Hopkins University, Baltimore, MD, USA
        {\tt\small (zzhan265@jhu.edu)}}%
\thanks{$^{3}$ Department of Neurology, The Johns Hopkins University School of Medicine, Baltimore, MD, USA
        {\tt\small (akherad1@jh.edu)}}
\thanks{$^{4}$ Department of Neuroscience, The Johns Hopkins University School of Medicine, Baltimore, MD, USA}
\thanks{$^{5}$ Department of Orthopedic Surgery, The Johns Hopkins University School of Medicine, Baltimore, MD, USA}
\thanks{$^{\dagger}$ Corresponding author}%
\thanks{$^{\ddagger}$ Equal contribution}%
}

\begin{document}

\maketitle
\thispagestyle{empty}
\pagestyle{empty}

\begin{abstract}
 

Realtime shape estimation of continuum objects and manipulators is essential for developing accurate planning and control paradigms. The existing methods that create dense point clouds from camera images, and/or use distinguishable markers on a deformable body have limitations in realtime tracking of large continuum objects/manipulators. The physical occlusion of markers can often compromise accurate shape estimation. We propose a robust method to estimate the shape of linear deformable objects in realtime using scattered and unordered key points. By utilizing a robust probability-based labeling algorithm, our approach identifies the true order of the detected key points and then reconstructs the shape using piecewise spline interpolation. The approach only relies on knowing the number of the key points and the interval between two neighboring points. We demonstrate the robustness of the method when key points are partially occluded. The proposed method is also integrated into a simulation in Unity for tracking the shape of a cable with a length of 1m and a radius of 5mm. The simulation results show that our proposed approach achieves an average length error of 1.07\% over the continuum's centerline and an average cross-section error of 2.11mm. The real-world experiments of tracking and estimating a heavy-load cable prove that the proposed approach is robust under occlusion and complex entanglement scenarios.

\end{abstract}

\section{INTRODUCTION}

In recent years, there has been an increased focus on shape estimation techniques in soft continuum robots \cite{manakov2021visual}, deformable object manipulation \cite{han2018robust}, minimally invasive surgery \cite{cabras2017adaptive, reilink3DPositionEstimation2013}, and industrial inspection \cite{ge2015cable}.
Researchers use the reconstructed shape to control the continuum manipulators \cite{croomVisualSensingContinuum2010,fan2020image}, estimate the trajectory of inserted flexible instruments \cite{shi2016shape}, and build digital twins for the soft tissues \cite{xu2021shape}.
These applications could be summarized as estimating the shape of deformable linear object (DLO) \cite{Choi2023}, or equivalently, finding the 3D centerline of the DLO \cite{joglekar2023suture}. DLOs are linear objects capable of elastic deformation, such as cables, wires, and rods \cite{caporali2023rt}.
However, accurate sensing and determining the shape of DLOs in realtime is challenging due to infinite degrees of freedom  \cite{zheng2022vise}, especially when the target is under the interference of complex surroundings and entanglements. 
This implies that for high-precision estimation, a large number of points on the continuum must be tracked in realtime, which significantly increases the computational complexity. 
Realtime processing is essential in some applications that involve continuum shape estimation. For example, force control of robotic arm with a continuum payload requires the dynamic modeling of the payload. This is particularly significant for our motivating clinical application, robot-assisted Transcranial Magnetic Stimulation (TMS).

\begin{figure}[t]
    \centering
    \includegraphics[width=0.8\linewidth]{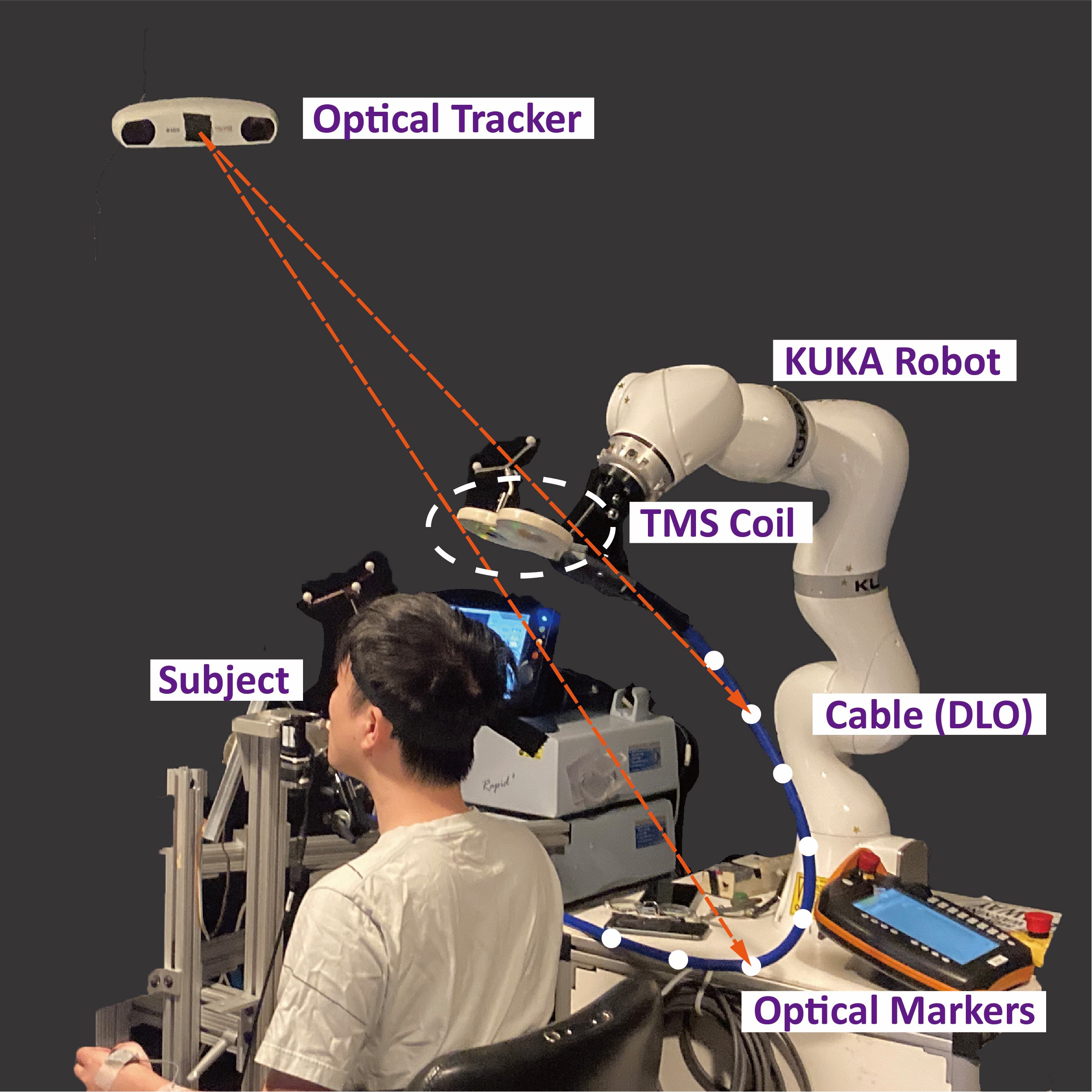}
    \caption{Experiment setup for robot-assisted TMS with optical tracking system (OTS) \cite{liu2022inside}. The cable is connected to the TMS coil and held by the robot. Optical markers highlighted with white circles are evenly placed on the cable. In robot-assisted TMS, estimating the shape of the cable is crucial for external force compensation and safety concerns. }
    \label{fig_setup}
   \end{figure}
   
 TMS is a noninvasive method to deliver electric stimulation to the brain \cite{zorn2011design}. Robot-assisted TMS employs robotic manipulators to increase the accuracy and safety of the TMS coil placement. An exemplary setup for robot-assisted TMS is shown in Fig \ref{fig_setup}. The force control of the robot can be adversely affected by the deformable attachments (coil cable) of the TMS, as the force applied to the robot depends on the shape of the deformable load. For instance, different configurations of the coil cable apply forces to the robot end-effector with different magnitudes and directions. Existing force control paradigms of robot-assisted TMS do not incorporate force control of the coil cable.

In this study, we develop a robust shape estimation pipeline to reconstruct the fixed-length DLO in realtime using optical tracker. When the number of the key points collected from the optical tracker is predefined, the proposed probabilistic point-labeling algorithm can successfully generate the sorting labels for the key points. The labels are used to rearrange the key points in the order they are placed along the actual continuum. The 3D centerline is reconstructed through piecewise $\mathcal{C}^2$ spline interpolation based on the labeled data. The pipeline is validated on both simulation and real DLO tracking tasks. The results show that our estimation is accurate and robust under occlusion and complex bending conditions, providing a solution to enhance force control of robot-assisted TMS and other soft robot tracking tasks.

\section{Related Work}

Different sensing methods are used for estimating the shape of a continuum in realtime, including fiber-optic sensors, RGB cameras, and infrared optical tracking systems (OTS). Many studies have integrated fiber-optic sensors into soft continuum robots to measure strain and develop dynamic models  \cite{shi2016shape, henken2012accuracy}.
These techniques are proposed for applications with small deformations, such as steerable needles for biopsy or ablation \cite{abushagur2014advances}. However, for continuum robots with large-scale and complex deformations, the application of these sensors may not be feasible.
This limitation has led to increased popularity of exteroceptive sensing approaches that are vision-based \cite{zheng2022vise} \cite{caporali2023deformable}. 

In contrast to the fiber-optic sensor method, vision-based approach uses visual sensors to detect the key points or the contour of the target and therefore it eliminates the impact on the design of the DLO. 
Several methods are articulated for detecting the shape based on 2D images with identifiable fiducials attached to the object \cite{manakov2021visual,hannanVisionBasedShape2003a}. 
The types of fiducials include LED lights with different intensities \cite{pedari2019spatial}, AR tags \cite{chang2020model}, and distinct reflective stickers \cite{hannanVisionBasedShape2003a}. 

Self-organizing map (SOM) \cite{kumar2004curve} is a possible alternative for fiducial markers. SOM requires collecting a dense point cloud from the target and it reconstructs the topological features according to the principle vectors generated from that point cloud. 
However, SOM cannot correctly detect the principle vectors if the point cloud is sparse.
The work in \cite{caporaliDeformableLinearObjects2023} proposed a stereo vision-based markerless shape-sensing algorithm for the concentric-tube continuum robot. 
Their work devised a unique SOM algorithm that iteratively adjusts the reference vector to approximate the backbone of the robot. 
Other estimation methods include incorporating 2D images with the strain basis function \cite{albeladi2021vision}, volumetric deformation model \cite{han2018robust}, and multi-view silhouette \cite{camarilloVisionBased3D2008}. 
The primary issue with these vision-based methods is that their environmental setup involves multiple cameras and a mono-colored background. Some of the working environments, like endoscopic surgery and robot-assisted TMS, are unable to meet the necessary conditions for this type of experimental setup, making them underperform in such complex scenarios. 

Compared to optical fiber and RGB-camera-based methods, OTS can capture the key point of the object with high accuracy and frequency, with fewer constraints on the continuum and the surroundings  \cite{asselin2018towards}.
However, OTS cannot track the order of each individual key point which is crucial for interpolation and reconstruction of DLO. In other words, the detected data points are unorganized/unordered sparse point clouds. Several sorting methodologies have been proposed in the past decades \cite{lee2000curve}.
Ge et al. \cite{ge2015cable} used OTS to capture the motion of a cable. They proposed a marker sorting algorithm that used Euclidean distance to identify the correct sequence of points and reconstructed the cable accordingly. However, this approach fails when multiple markers appear within a certain proximity. Our approach presented here has been evaluated in comparison to the approach proposed by Ge et al.

Key point labeling is required for shape estimation once a point cloud is obtained from sensors. Labeling algorithms solve this problem by determining the order of the points \cite{alexanderson2017real}. Labeling algorithms are broadly used in the motion capture process where unordered point clouds are employed to reconstruct the true motion of the target \cite{bascones2019robust}. Prior approaches for labeling typically view the target object as a system composed of multiple rigid bodies and address the connections among the bodies as kinematic constraints. Several algorithms have been applied to establish a stable and optimal labeling method, including genetic algorithm and tree-search algorithm \cite{bascones2019robust}. However, few studies presented a robust and stable labeling algorithm based on the geometrical properties of DLOs while applying realtime labeling processes to estimate the actual shape of the DLO.

\section{Method}

Our proposed shape estimation pipeline can be decomposed into key point labeling, interpolation, and optimization. The primary objective of this work is to focus on key point labeling. We propose a novel labeling algorithm called Probabilistic Continuum Key Point Labeling Graph (PCLG) to determine the order of key points along the continuum. PCLG maps out the scattered key points and identifies potential neighboring points. 

\begin{figure}[thpb]
    \centering
    \includegraphics[width=0.45\textwidth]{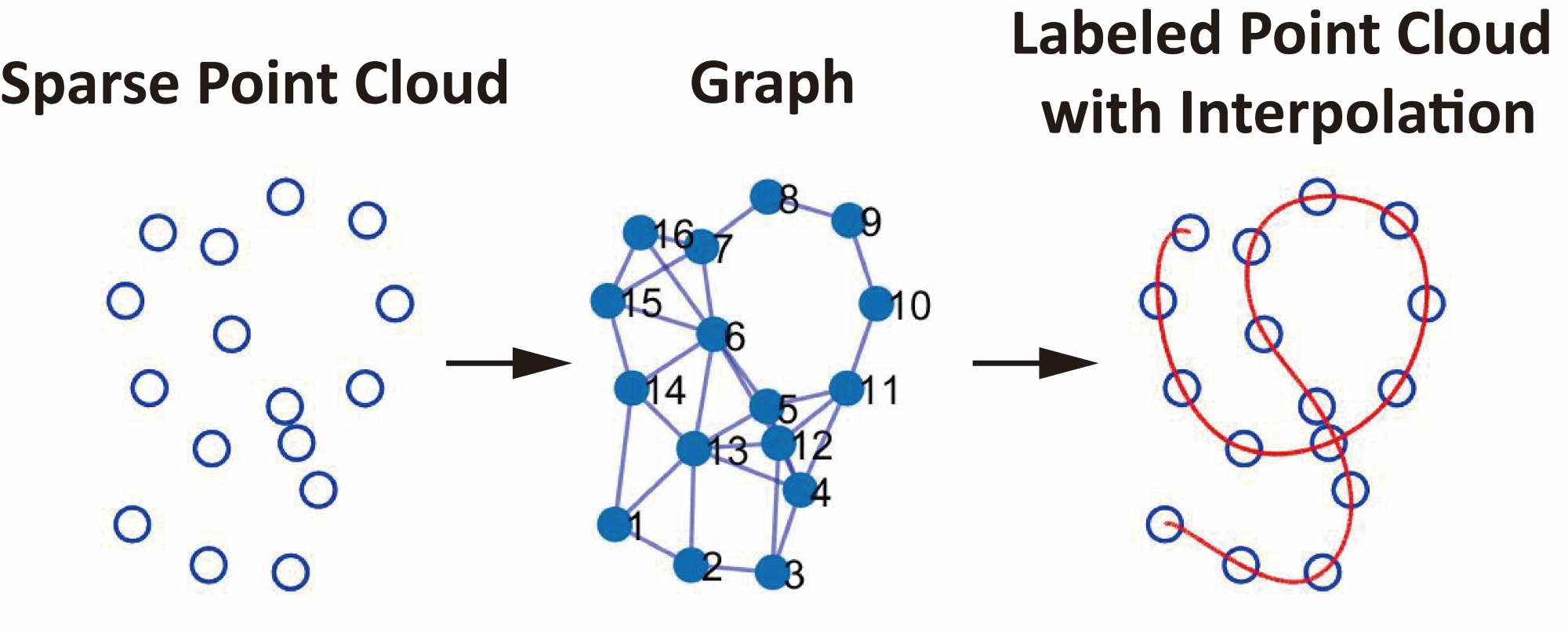}
    \caption{Simulated DLO with nodes highlighted. The nodes are represented with circles in all three panels. A weighted graph in the middle panel is used to describe the connectivity of the nodes. The nodes are located at the same intervals along the center line of the DLO. The graph edges are initialized based on the Euclidean distances between the nodes. The right panel depicts the labeled nodes after pruning the edges.}
    \label{fig map}
   \end{figure}

In PCLG, the key points are stored in a graph data structure. Each node represents a key point, and the potential candidates of the neighboring points are connected using the edges of the graph.
The weights assigned to the edges of the PCLG are determined by the likelihood of the corresponding points being neighbors (Section \ref{sec:graph}). This labeling result is then used to perform piecewise 3D interpolation to connect the points. This process can be summarized as Fig. \ref{fig map}. By assuming that the geometric information (the number and distribution of the key points on real DLO)  is known beforehand, the interpolated curve is fine-tuned by optimizing the total length and strain of the DLO.

\subsection{Key Point Labeling Formulation}
   
Shape estimation is the method to find the topological relations among points \cite{kumar2004curve}, based on the geometrical or dynamic properties of the object. The actual shape of the DLO could be precisely reconstructed connecting the key point with labeled order. To build a robust and computationally efficient model to estimate the shape of the DLO, we use OTS to localize the retro-reflective markers attached to the continuum object and sort the markers to reflect the shape of the continuum object.

Let $P(s) = [p_1(s), p_2(s), ..., p_i(s), ... ]$ be an array where the points are sequentially selected from the beginning to the end of $s$, which denotes the centerline of a DLO. $O(s) = [..., p_i^j(s), ... ]$ is the point array derived from optical tracker, where $p_i^j(s)$ represents the points that are at the $i$-th place of $P(s)$ but at $j$-th place in $O(s)$. As shown in Eq. \ref{eq_permutation}, labeling is equivalent to finding the permutation matrix $\mathcal{M}_p$ that changes the order of the points in $O(s)$ to match $P(s)$.

\begin{equation}
    P(s) = \mathcal{M}_p O(s)
    \label{eq_permutation}
\end{equation}

\subsection{PCLG Initialization}

The points on the DLO are initialized as an unweighted graph. Each node $ p_m \in \mathbb{R}^3 $ represents one point and the edges connecting the nodes stand for the possibility of connectivity. 
When the edges are initialized, the graph is connected based on the Euclidean distance between the nodes. If the segment of the continuum object is measured $\delta_L$, the connectivity between node $p_m$ and $p_n$ is set as:

\begin{equation}
    \mathcal{C}(e_{m,n}) =  \begin{cases} 1, & \mbox{if } ||p_n - p_m|| \leq \delta_L + \sigma_L \\ 
    0, & \mbox{if } ||p_n - p_m|| > \delta_L + \sigma_L
    \end{cases}
\end{equation}

where $\sigma_L$ is a parameter that reflects the reliability of the sensors. The middle panel in Fig. \ref{fig map} demonstrates the graph after initialization. 
The nodes are assumed to be evenly placed on the DLO with a distance of $\delta_L$ along the principle axis (center line). Only the nodes located within $\delta_L + \sigma_L$ of each other will be connected by edges. A continuum shape requires each non-terminal node to have only two neighbors. Generally, redundant edges (3 or more edges at a non-terminal node) remain after edge initialization. The redundancy indicates that those node pairs connected by multiple edges are potential candidates of the actual neighboring nodes pair. Therefore, edge pruning process is required to select the edges that are most likely to reflect the actual shape and ensure that every non-terminal node has only two neighbors. 

\subsection{Probabilistic Edge Pruning} \label{sec:graph}

To cut down the redundant edges, a probabilistic distribution algorithm is implemented to assign a weight to each edge, representing the likelihood of that edge being a valid solution. The algorithm is based on approximating the DLO as a piecewise constant curvature (PCC) model \cite{hannanVisionBasedShape2003a}. PCC model assumes the DLO/continuum robot as a series of mutually tangent constant-curvature arcs \cite{webster2010design}. Based on PCC hypothesis, we can determine whether an edge is a valid solution by assessing the probability of the two points on the edge being linked by an arc with a fixed length.

\begin{figure}[t]
    \centering
    \includegraphics[width=0.85\linewidth]{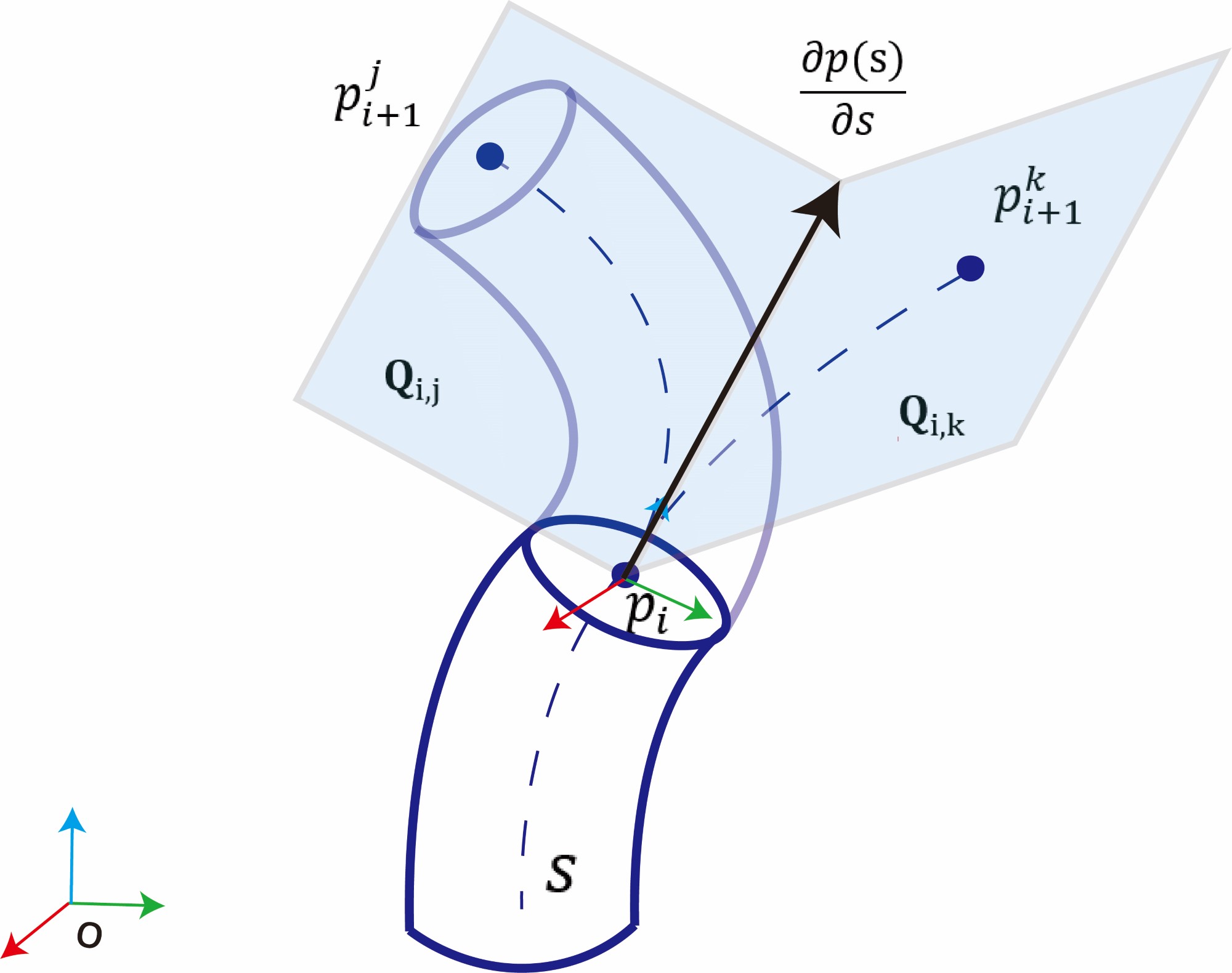}
    \caption{An illustration of two subsequent segments of a DLO $s$. Two segments are tangential at $p_i$ and the tangent vector is $\frac{\partial p}{\partial s}$. $p_{i+1}^j$ and $p_{i+1}^k$ are potential adjacent nodes with $p_i$. Plane $\bf{Q}_{i,j}$ is defined by the points $p_{i+1}^j$, $p_i$, and $\frac{\partial p}{\partial s}$; Plane $\bf{Q}_{i,k}$ is defined by the points $p_{i+1}^k$, $p_i$, and $\frac{\partial p}{\partial s}$. The dashed line represents the center line of the DLO. }
    \label{fig3d}
\end{figure}

As shown in Fig. \ref{fig3d}, $p$ denotes the point on a DLO $\mathbf{s}$, and the tangent vector at $p$ is $\frac{\partial p }{\partial s}$. Consider a sequence of points where the $i$-th point is denoted by $p_i$. Let $p_{i+1}^j$ and $p_{i+1}^k$ be two points connected to $p_i$ by edges. The arc connecting $p_i$ and $p_{i+1}^j$ lies within the plane (denoted as $\bf{Q}_{i,j}$) that is formed by these two points and cross-section normal $\frac{\partial p }{\partial s}$. Therefore, the task becomes to find a suitable arc on $\bf{Q}_{i,j}$, which converts a three-dimensional problem into a two-dimensional planar problem.

\begin{figure*}
    \centering
    {\includegraphics[width=0.75\linewidth]{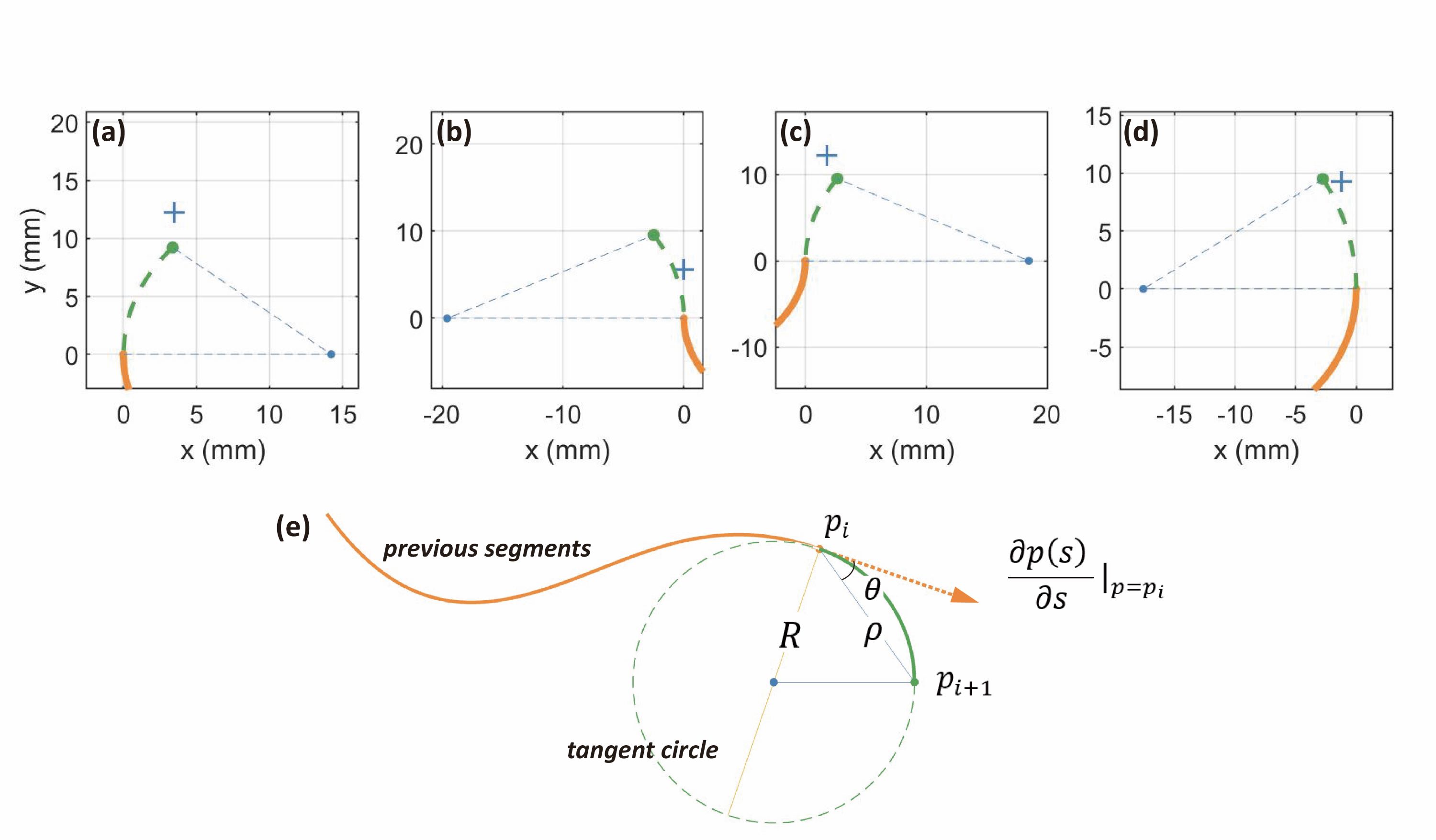}}
    \caption{(a)-(d) are four examples that illustrate the noisy measured points versus the simulated ground truth of the DLO segments. Blue crosses are the noisy measured value $\tilde{p}_{i+1}^j$. Green dots represent the ideal value $p_{i+1}^j$. Orange dots are the base point $p_i$. The ideal arc connecting $p_i$ and $p_{i+1}^j$ is represented by green dashed lines. The orange lines are the previous segment. (e) depicts how the tangent circle $\mathbf{c}$ and the current segment are defined on a DLO. The center of the $\mathbf{c}$ is the blue dot in (a)-(e). Blue dashed lines define the radius of $\mathbf{c}$.}
    \label{fig_theory}
   \end{figure*}

Letting the first point $p_i$ be the origin and construct a polar coordinate system accordingly on the plane $\bf{Q}_{i,j}$, we can express the position of the next point $p_{i+1}^j$ as $p_{i+1}^j = [\rho_{i+1},\theta_{i+1}]^T$. Fig. \ref{fig_theory}(e) demonstrates the definition of the parameters. With the length of the arc between $p_i$ and $p_{i+1}^j$ being $\delta_L$, the radius of the arc is expressed as $R = \delta_L / (2\theta)$. Eq. \ref{eq_sine} describes the possible positions of $p_{i+1}^j$ with respect to $p_i$. 

\begin{equation}
        \rho_{i+1} = \frac{R \sin{(2\theta_{i+1}})}{\sin{(\pi/2 - \theta_{i+1})}}  \Rightarrow \rho_{i+1} = \frac{\delta_L \sin{(\theta_{i+1})}}{\theta_{i+1}}
        \label{eq_sine}
\end{equation}

Notably, a constraint exists as $\theta \in [0, \pi]$. By adjusting the value of $ \theta $, it is possible to generate all possible positions of the next point.

Additionally, due to imperfections of the sensor performance, the detected location $\tilde{p}_{i+1}^j$ may be situated in the neighboring area of $p_{i+1}^j$, as illustrated in Fig. \ref{fig_theory} (a)-(d). Considering the sensory noise and other perturbations as random factors, we establish a scalar field of probability distribution based on the trajectory defined by Eq. \ref{eq_sine}. The probability of every point in the scalar field represents the likelihood of that point being $p_{i+1}$. The field is constructed by assuming the measured coordinates $\tilde p_{i+1}$ is subject to some Gaussian distribution $\mathcal{N}(p;\mu, \Sigma)$, where $\mu = [\mu_{\rho}, \mu_{\theta}]^T$. $\mu_{\theta} $ ranges from 0 to $\pi$ and $\mu_{\rho}$ is evaluated by substituting the value of $\mu_{\theta}$ in Eq. \ref{eq_sine}. The coordinate of measurement is computed by:

\begin{equation}
\begin{cases}
        \theta = \arccos(\frac{\vec n \cdot (\tilde p_{i+1} - p_i)}{||\tilde p_{i+1} - p_i||}) \\
        \rho = ||\tilde p_{i+1} - p_i||
\end{cases}
\label{eq_angle}
\end{equation}

Therefore, the probability of any point in the field being the correct neighboring point is expressed as: 

\begin{equation}
    P( p_{i+1} = p_{i+1}^j | p_i) = \mathcal{N}( \tilde p_{i+1}^j; \mu, \Sigma)
    \label{eq_field}
\end{equation}

\begin{figure}[t]
    \centering
    \includegraphics[width=0.85\linewidth]{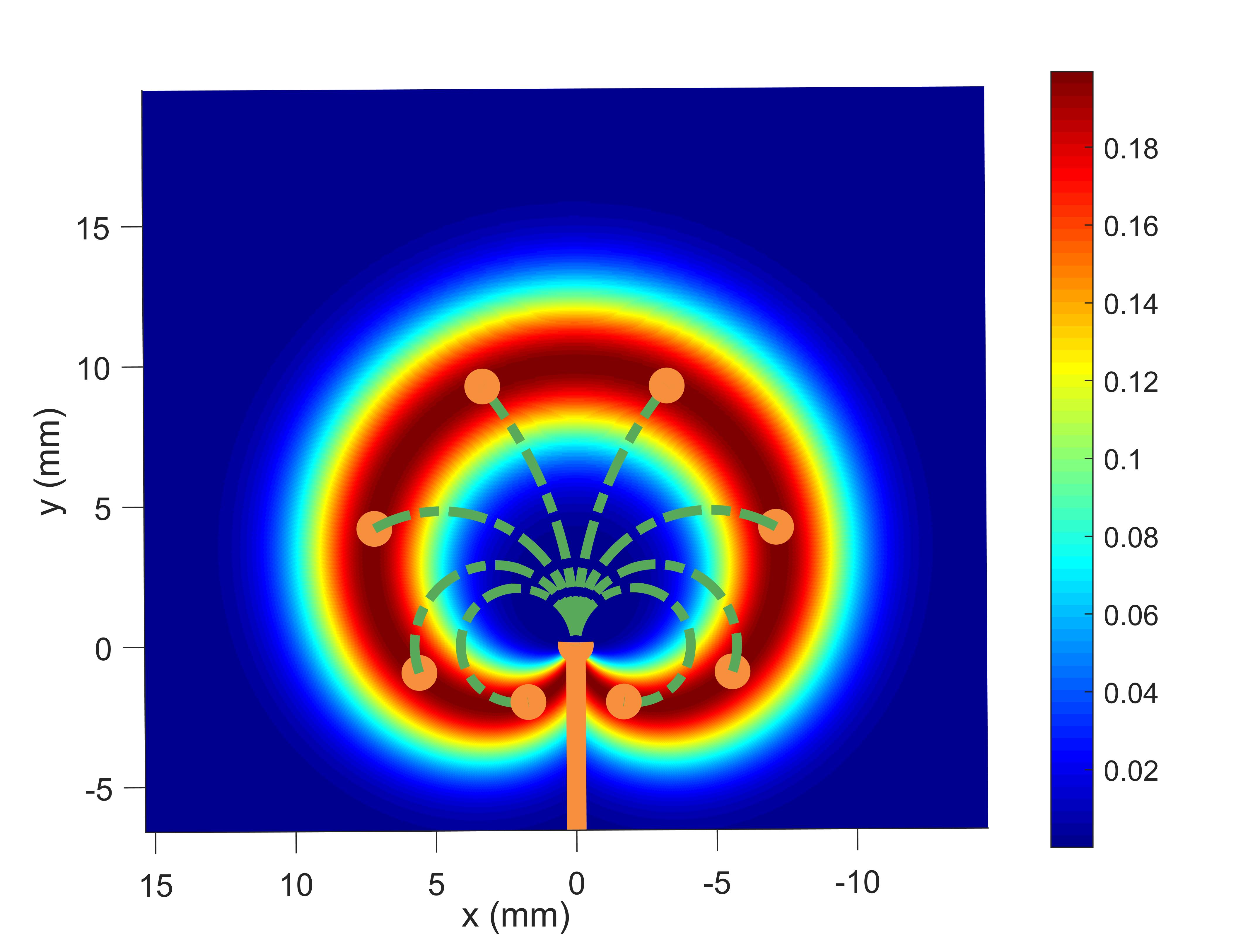}
    \caption{Eight possible configurations are overlaid on the planar probability distribution field, which evaluates the likelihood of a node being the next adjacent node. The preset distance between nodes is 10mm in this example. The nodes are shown with orange dots. Arcs that connect the adjacent nodes are plotted as green dashed lines. The orange solid line denotes the previous segment. }
    \label{fig prob2d}
\end{figure}

Note that the probability is assigned to $\mathcal{C}(e_{i,j})$ to represent the weights of the edges. Fig. \ref{fig prob2d} shows the probability field of a key point and some possible positions of the consecutive nodes. The scalar field for all key points can be established, except for the terminal key points. Every time the map is generated, a searching algorithm will parse all edges to evaluate the weights by querying the probability field. The edge with the highest probability connects the nodes that are most likely to be adjacent to each other. The pseudocode of this process is presented in Algorithm \ref{alg:cap}. The algorithm produces a list of labeled key points, referred to as $P_{open}$.

This labeling algorithm can be articulated as: Given a set of points $O(s) = [p_0, p_1, \ldots, p_n]$ with random order, if the order up to $k-1, k \in [0, \ldots ,n]$ is determined, the $k$-th point  $p_k$ can be found by evaluating the rest of the points in the probabilistic field for $p_{k-1}$ and select the one $p_k$ that maximizes the probability. The non-terminal nodes in the PCLG are connected with only two other nodes by pruning the redundant edges with lower probabilities. By utilizing this labeling method, we can determine the permutation matrix $ \mathcal{M}_p $ and retrieve the node sequence $P(s)$. Subsequently, $P(s)$  is applied to reconstruct the shape.

\begin{algorithm}
\caption{PCLG Algorithm}\label{alg:cap}
\begin{algorithmic}
\Require 3D points in random order stored in $P_{random}$
\Require $P_{open}$ the open set that contains the sorted points
\State $\sigma \gets $ standard deviation of the measurement
\State $P \gets$ initialize as the probability of the edges
\State $F \gets$ the probability distribution field, defined in Eq.\ref{eq_field}
\State $N \gets $ length($P_{random}$)
\State $G$.init\_graph\_from\_points($P_{random}$)
\For{node in G}

\For{edge in G that contains this node}

$[\rho, \theta] \gets$ get\_connected\_node(edge)

$P$.pushback($F(\rho, \theta, \sigma)$)

\EndFor

Next\_Index = $max(P)$

G.edges(edge\_idx $\neq$ Next\_Index) = null 

\EndFor

\State $p_c \gets$ get\_endpoint\_of\_edge(edge)

\State $P_{open}$.pushback($p_c$)
\For{node in $G$}

$p_{next} \gets$ get\_adjacent\_node($p$)

$P_{open}$.pushback($p_{next})$

\EndFor
\end{algorithmic}
\end{algorithm}

To avoid erroneous labeling and enhance robustness, we incorporate a minimal stress optimization.
The total stress of a DLO is computed through $\sigma = (\mathbf{m} \times \mathbf{y})/\mathcal{I}$, where $\sigma$ is the bending stress, $\mathbf{m} $ is the moment applied, and $\mathbf{y}$ is the distance from the neutral axis to the point where the stress is calculated.
Given that the moment stays the same at a specific state, the bending stress can reach a minimal value when the distance $\mathbf{y}$ is minimized. 
The optimization process eliminates cases where there are two edges with comparable probabilities and guarantees that $P(s)$ always represents the shape with the least strain.

\section{Experiments and Results}
\label{sec result}

The primary experimental setups are shown in Fig. \ref{fig_setup} and \ref{figRGB}. Details of the TMS infrastructure can be found in our previous work in \cite{liu2022inside, liu2023toward, liu2023GBEC}. We use the NDI Polaris Vicra (Northern Digital Inc., Waterloo, ON, Canada) to track the retro-reflective markers and a RealSense (Intel, Santa Clara, CA, USA) camera to help visualize the actual shape. Reconstruction accuracy is evaluated under simulation and controlled experiment setups.

\subsection{Labeling Error Evaluation}

The labeling error is evaluated in a cable tracking task. The ground truth order is obtained through manual localization using a probe with its tip position being tracked by Polaris. To get the ground truth, the tracked probe is used to touch one of the key points on the cable for a duration of 20 seconds. The actual order of the touched point $\hat{p}_i$ is determined by visual inspection. Thus, we can compare the order derived from the labeling algorithm and the actual order of that particular point, i.e. $||\hat{p}_i - p_i|| $ across the period. 

\begin{figure}[t]
    \centering
    \includegraphics[width=0.8\linewidth]{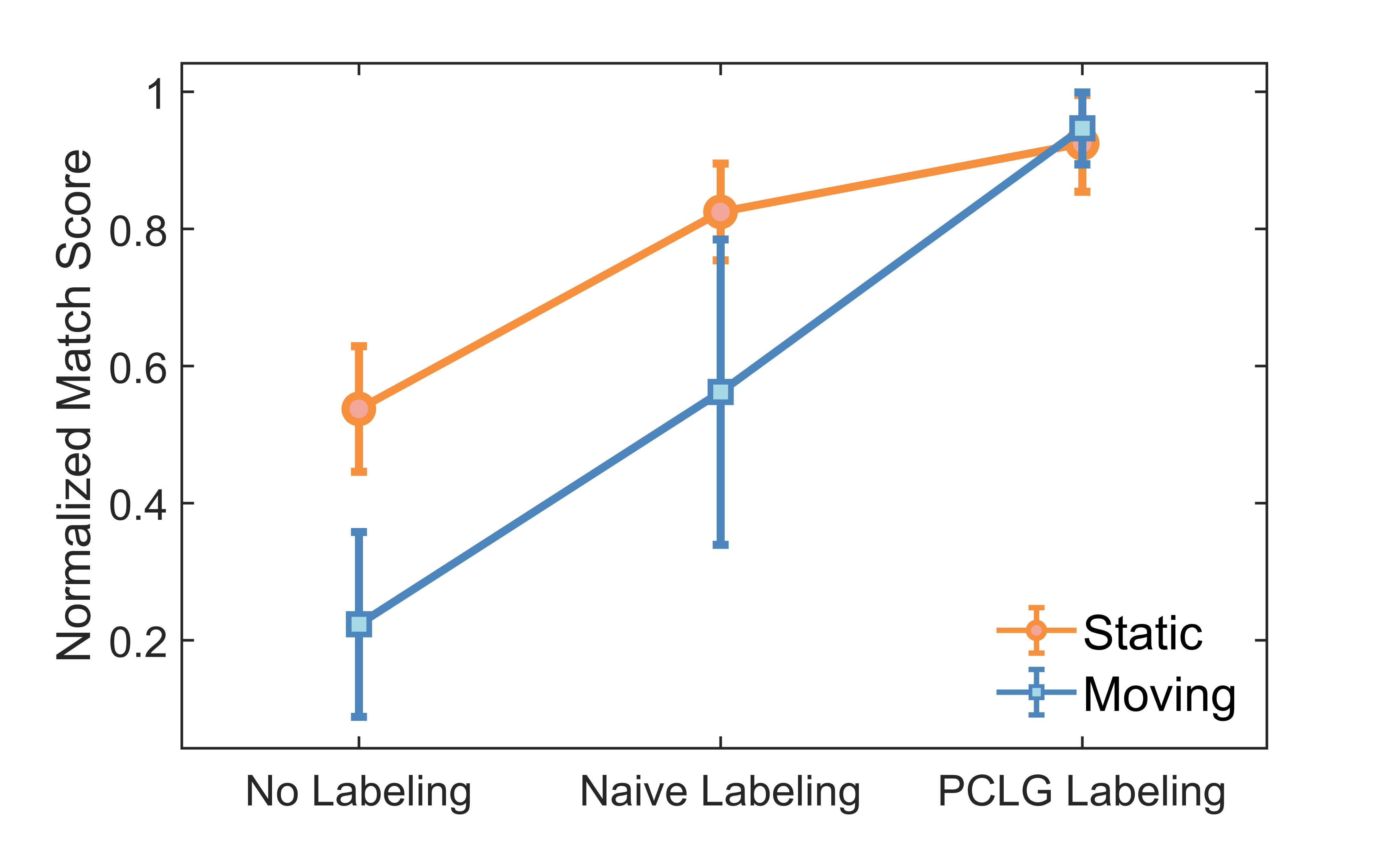}
    \caption{Comparison of the performance of different labeling methods for both object-static and object-moving cases. The performance is evaluated by normalized match score $M$. The range of the vertical axis is from 0 to 1. ``Static'' refers to the cable being fixed whereas ``moving'' refers to the cable being attached to the robot end-effector and moving along with the robot.}
    \label{fig_error}
\end{figure}

A boolean parameter $m(t)$ describes whether the labeling result correctly matches the ground truth at time $t$. If the Euclidean distance between the labeled point and the true position is within 2 mm, $m$ will be assigned as "1", indicating a correct match is found. A normalized match score $M = \frac{1}{t_1 - t_0} \sum_{t_0}^{t_1} m(t)$ is used to reflect the overall success rate, with a maximum value of 1. The normalized match score for three different labeling methods under both object moving and static situations is depicted in Fig. \ref{fig_error}. The labeling methods are No Labeling, Naive Labeling \cite{ge2015cable}, and PCLG Labeling. The Naive Labeling method, which is typically used for DLO labeling tasks, adopts a maximum distance constraint that only connects the nearest point pairs. Although in the static situation, the Naive Labeling method produces quite satisfactory $M$ of 0.825 $\pm$ 0.077, PCLG reduces the variation and therefore gains better performance in terms of robustness. On average, PCLG labeling achieves a correct rate of 94.7\% for the object-moving task. The performance of PCLG was significantly better than the naive method, resulting in a 68.5\% increase in the correct rate, and a 76.4\% decrease in the standard deviation.

\subsection{RGBD Image Comparison}

\begin{figure*}
    \centering
    {\includegraphics[width=\linewidth]{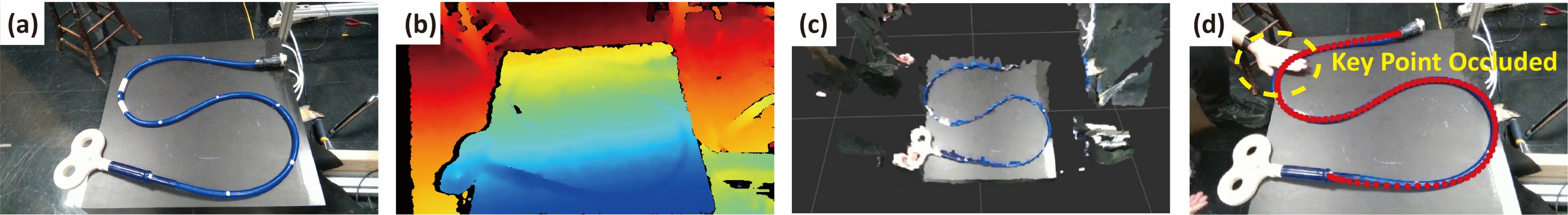}}
    \caption{(a)-(c) are the RGB image, depth image, and point cloud of the DLO, respectively. (d) is the estimation results overlaid on the RGB image. The red dots in (d) are the interpolation results. 
 One key point is blocked but the estimation result still matches the actual shape. }
    \label{figRGB}
   \end{figure*}

An RGBD camera is used to validate the reconstruction result by evaluating the alignment error. The RGB image collected from RealSense is shown in Fig. \ref{figRGB}(a). The points obtained from NDI (denoted as $\mathbf{p}_{p}$) are projected onto the image plane through the camera-tracker transformation  $^t T_c$ (calibrated using  \cite{asselin2018towards}) and camera intrinsic transformation $\mathcal{K}$. The projection homogeneous coordinates $\mathbf{p'_p} = [x'_p,y'_p,1]^T$ are solved by:

\begin{equation}
\mathbf{p'_p} = \lambda \begin{bmatrix}
    \mathcal{K} & \mathbf{0}_{3\times1}
    \end{bmatrix}
    \begin{bmatrix}
        ^t\mathbf{R}_c & ^tp_c \\ \mathbf{0}_{1 \times 3} & 1
    \end{bmatrix} 
    \begin{bmatrix}
        p_{p} \\ 1
    \end{bmatrix}
\end{equation}

To identify alignment errors introduced by the camera-tracker calibration, the errors of both calibration and the projection are calculated and summarized in Table \ref{table:mse}. Calibration errors in \textit{Cartesian} space ($\epsilon_c$) and \textit{pixel} coordinates ($\epsilon_p$) are computed by localizing the same point (identified by an ArUco \cite{garrido2014automatic} tag) in two sensors. We only measure the pixel differences of the key points and use them as projection errors since projecting the RGB image back to the 3D space accurately is challenging due to the lack of depth information.

\begin{table}[ht]
\caption{Alignment Errors of Camera-Tracker Calibration and Projection Result}

\centering      
\begin{tabular}{c c c }  
\hline\hline
&& \\ [-1ex]
\hspace{1pt} & $\epsilon_c$ & $\epsilon_p$ \\ [1ex] 
\hline           
&& \\ [-1ex]

Calibration & 2.95mm $\pm$ 0.41mm & 3.33px $\pm$ 1.08px \\ [1ex]    
Projection & - &  4.02px $\pm$ 1.49px\\ [1ex]
Projection when blocked & - & 9.65px $\pm$ 3.50px \\ [1ex]
\hline     
\end{tabular} \label{table:mse} 
\end{table}

Our method improves the robustness of the estimation process as shown in Fig. \ref{figRGB}(d). Compared with \cite{ge2015cable}, the pixel error of PCLG reconstruction did not significantly increase when some markers were occluded, whereas the naive approach often fails in such cases. With limited occlusion, our method outperforms the Coherent Point Drift method reported in \cite{chi2019occlusion}.

\subsection{Simulation}

\begin{figure}[t]
    \centering
    \includegraphics[width=0.85\linewidth]{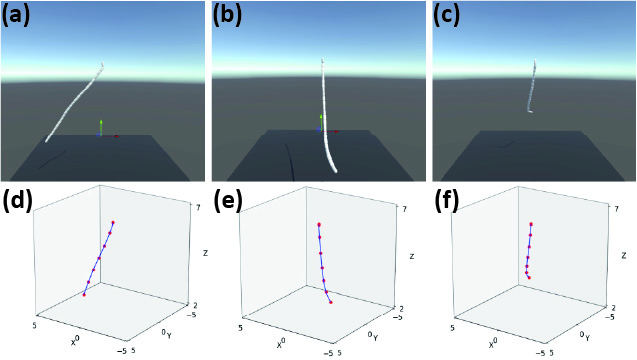}
    \caption{Comparison of the simulated model and its reconstruction. The simulated model consisted of 128 spring-mass segments, among which the first segment was fixed and the rest were only connected at the end without global fixation. We selected 7 endpoints of the segments and randomized their order to mimic the behavior of the real OTS. (a)-(c) are snapshots taken while the model is being moved under manually specified impulse applied on the end. (d)-(e) are the corresponding reconstructions derived by PCLG sorting and spline interpolation. }
    \label{fig_simulation}
\end{figure}

To validate the robustness and accuracy of our method, Unity was employed to construct a simulation model that serves as a benchmark for our tests. We computed the Euclidean distance between corresponding points on both datasets to gauge the discrepancies. An example of the simulated DLO and the estimated DLO is shown in Fig. \ref{fig_simulation}.

The data consisted of 128 scattered points evenly distributed along the simulated spring-mass model, with a radius of 5mm and a total length of 1m. We selected 8, 16, and 32 points separately out of a total of 128 by evenly spacing them and randomly shuffling their order to simulate the readings from the optical tracker. Our algorithm used the selected points to sort the points and interpolated the curves in between. 
The results were then compared with the ground truth model regarding the total length errors $\epsilon_l$ and average cross-section deviation errors $\epsilon_d$.

The results from these comparisons are presented in Table \ref{table_unity}. The percentage of total length errors was recorded instead of absolute values. Our algorithm gains better results compared to the point cloud-based method reported in the literature \cite{schulman2013tracking}. It can be observed that overall errors decrease as the number of nodes increases. This is due to the fact that fewer nodes may not be able to capture all the curvature variations and therefore add up to the total length error.

\begin{table}[ht]
\caption{Percentage Errors for the Total Length and the Absolute Cross-section Deviation Errors in 3D Space}

\centering      
\begin{tabular}{c c c c c}  
\hline
\hline
&& \\ [-1ex]
Nodes \# & $\mu(\epsilon_l) \pm \sigma(\epsilon_l)$ & $\max(\epsilon_l)$ & $\mu(\epsilon_d) \pm \sigma(\epsilon_d)$ & $\max(\epsilon_d)$\\ [1ex] 
\hline           
&& \\ [-1ex]

8 & 1.30  $\pm$ 0.47\% & 2.45\% & 2.19 $\pm$ 1.08mm& 4.86mm\\ [1ex]    
16 &  0.99 $\pm$ 0.34\% & 1.55\% & 2.01 $\pm$ 0.99mm& 3.73mm\\ [1ex]       
32 &  \textbf{0.92} $\pm$ \textbf{0.35\%} & \textbf{1.49\%} & \textbf{2.13} $\pm$\textbf{ 0.51mm}& \textbf{3.44mm}\\ [1ex]       
\hline     
\end{tabular} \label{table_unity} 
\end{table}

\section{CONCLUSIONS AND FUTURE WORK}

This study focuses on tracking and reconstructing the 3D shape of a deformable linear object in realtime through optical tracking system, which can be applied to localize soft robots or establish digital twins for cables. 
Our method is applicable in situations where the key points are distributed evenly along the target and where the number of the key points is known beforehand. 
The results show that our method is capable of shape reconstruction in complexities such as cable entanglement and occlusions. 

Nevertheless, the algorithm fails to distinguish the points when multiple markers are colinear with the line of sight of the optical tracker, which could lead to a completely incorrect estimation. Moreover, our current algorithm doesn't consider temporal information as a consistency constraint during motion. In the future, by taking the optic flows \cite{shah2021traditional,revathi2012certain} as a frame-to-frame constraint to optimize the result, we could potentially improve the stability, especially in the case of sudden motion and temporary occlusion.

\bibliographystyle{IEEEtran}
\bibliography{ref}

\end{document}